\documentclass[conference]{IEEEtran}
\IEEEoverridecommandlockouts
\usepackage{cite}
\usepackage{amsmath,amssymb,amsfonts}
\usepackage{algorithmic}
\usepackage{graphicx}
\usepackage{textcomp}
\usepackage{xcolor}
\usepackage{authblk}
\usepackage[utf8]{inputenc}
\usepackage[T5]{fontenc}
\usepackage{hyperref}
\usepackage{multirow}
\usepackage{float}
\usepackage{makecell}
\usepackage{booktabs}
\usepackage{subcaption}

\def\BibTeX{{\rm B\kern-.05em{\sc i\kern-.025em b}\kern-.08em
    T\kern-.1667em\lower.7ex\hbox{E}\kern-.125emX}}
\begin{document}

\title{ViTOED: A Dataset for Target-Oriented Emotion Detection on Vietnamese Social Media Texts
}

\author[1,2]{Chanh Vo}
\author[1,2]{Son T. Luu}
\author[1,2,*]{Ngan Luu-Thuy Nguyen\thanks{*Corresponding author}}
\affil[1]{Faculty of Information Science and Engineering, University of Information Technology, Ho Chi Minh City}
\affil[2]{Vietnam National University, Ho Chi Minh City, Vietnam}
\affil[ ]{Email: \textit {20521122@gm.uit.edu.vn}, \textit{sonlt@uit.edu.vn}, \textit{ngannlt@uit.edu.vn}}

\maketitle

\begin{abstract}
This paper introduces ViTOED, a novel dataset for target-oriented emotion detection in Vietnamese social media texts. The ViTOED comprises 10,985 user comments and 21,244 manually annotated opinion quadruples (source, target, expression, polarity) that follow strict guidelines. The dataset reveals Vietnamese-specific phenomena, such as implicit sources and targets and vocabulary ambiguities, enabling deeper analysis of user emotions toward entities. We propose a baseline using structured sentiment graphs and evaluate various Vietnamese pre-trained language models. The empirical results highlight challenges in span detection and relation extraction and indicate substantial room for model improvement in Vietnamese Target-Oriented Emotion Detection tasks.
\end{abstract}


\section{Introduction}
\label{introduction}
Sentiment Analysis (SA) is a task in Natural Language Processing (NLP) that analyzes users' opinions and emotions toward specific entities \cite{liu2022sentiment}. In sentiment analysis, detecting Entity and Aspect levels in users' input text is a challenging but potentially useful task for performing qualitative and quantitative analyses of users' emotions toward specific targets. According to the definition from \cite{liu2022sentiment}, "an actual word or phrase indicating an entity category is called the expression". One of the challenges for sentiment analysis is to detect the "implicit expressions", which contain no polarity markers but still carry the human-aware polarity in context for expressing the emotions \cite{li-etal-2021-learning-implicit}. Moreover, a user's text can contain multiple entities, and each entity can have its own expression to a specific target entity with corresponding polarity. Exploiting the emotion associated with a target helps understand the rationale for the polarity carried by individuals. It improves the accuracy of sentiment analysis of users' social media text. 

\begin{figure}
    \centering
    \resizebox{.5\textwidth}{!}{  
        \includegraphics[width=\textwidth]{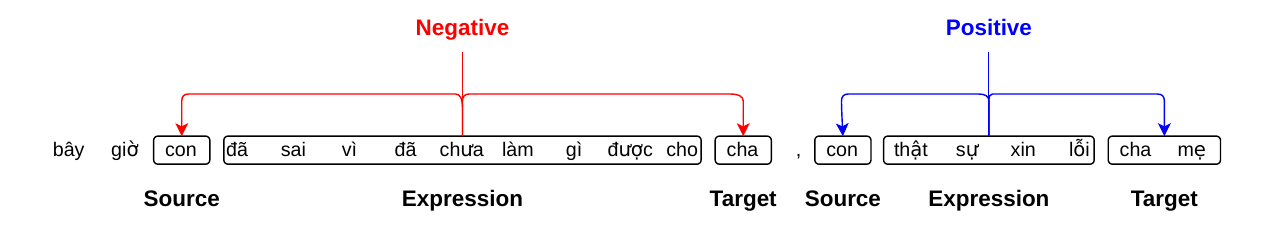}
    }
    \caption{A sample of target-oriented emotion detection from the sentence (English: I was wrong for not being able to do anything for my father. I am truly sorry to my parents.)}
    \label{fig_sample}
\end{figure}

To be clear, the task of target-oriented emotion detection is denoted as giving a sentence $s$ in human language, the goal is detecting a list of opinion tuples $O=\{o_{1}, o_{2}, ..., o_{n}\}$, in which each $o \in O$ is a quadruple of \textit{<source, target, expression, polarity>}. The source indicates the holder of expression in the sentence, and the target is the subject to which the source refers. The expression refers to the emotions or behaviors expressed by users. Finally, the polarity indicates the sentiment or opinion of the users. Figure \ref{fig_sample} illustrates an example for detecting the emotion by the target in the user's sentence. At first, this sentence is categorized as "Sadness" (as reported in the UIT-VSMEC dataset \cite{ho2020emotion}), but it actually expresses two opinions. One represents the "negative" emotion between the source "con" (in the Vietnamese language, this is the pronunciation between son and daughter with his/her parent) toward the target "cha" (English:  father) through the expression "đã sai vì chưa làm được gì cho" (English: fault because he/she disobeys his/her father). The remaining emotion expresses the emotion of the source "con" with the target "cha mẹ " (English: parents) as positive through the expression "thật sự xin lỗi" (English: hereby apologize). This is positive because it shows that the subject (the source) in the sentence recognizes their mistake and apologizes to their parents, indicating empathy toward the target.

To address the problem of target-oriented emotion detection in Vietnamese, we proposed the ViTOED dataset, which contains about 10,985 sentences collected from Vietnamese social networking sites. There are 21,244 manually annotated opinion quadruples in which each sentence can have multiple opinions. Besides, we construct a baseline for target-oriented sentiment analysis based on \cite{barnes-etal-2021-structured} and evaluate various Vietnamese pre-trained language models on our dataset to assess the ability of computers to understand the emotions expressed toward a specific target by the source. To sum up, our paper has three main contributions: (1) We introduce ViTOED, a new human-annotated Vietnamese dataset for emotion detection by target. The dataset is constructed under a strict annotation process, (2) We analyze the dataset to identify a phenomenon in Vietnamese regarding the difference between the target, source, and expression in expressing emotion in a sentence, and (3) We evaluate various Vietnamese pre-trained language models using the Structured Sentiment Analysis Graph proposed by \cite{barnes-etal-2021-structured} on our dataset to investigate the performance of machine learning models for target-oriented emotion tasks in Vietnamese.  

\section{Related Works}
\label{relatedworks}
Although Sentiment analysis (SA) plays a vital role in many real-world applications, this task has several critical challenges. According to \cite{venkit-etal-2023-sentiment}, biases and harms in sentiment analysis models make SA models unable to represent all users' perspectives in the real world. This bias issue is due to the lack of explanation in predicting the sentiment through "an interdisciplinary lens" \cite{venkit-etal-2023-sentiment}. To reduce bias, SA models need to capture external factors that contribute to sentiment rather than focusing solely on lexical text categorized as positive or negative. One potential approach to reducing bias in the SA is \textit{targeted sentiment analysis} \cite{barnes-etal-2021-structured}, in which SA models exploit targets, holders, expressions, and their polarity to form an argumentative opinion about specific objects in the sentence. There are several datasets employed for this task, such as the $DS_{Unis}$ \cite{toprak-etal-2010-sentence} in English and MultiBooked \cite{barnes-etal-2018-multibooked} in Spanish. These available datasets are valuable for building and evaluating the target-oriented sentiment models.

In Vietnamese, most large-scale datasets currently available focus on the Aspect-based Sentiment Analysis (ABSA) task \cite{10.1145/3610226}. The high-quality, human-annotated datasets available in ABSA provide a valuable resource for constructing and evaluating ABSA models. Some of the ABSA datasets are the VLSP2018 \cite{nguyen2018vlsp} and UIT-ViSFD \cite{luc2021sa2sl}. Those ABSA datasets are constructed at the sentence level, making it difficult to exploit the polarity represented by a set of tokens or words within a sentence. Additionally, the UIT-ViSD4SA dataset \cite{thanh-etal-2021-span} enables computers to perform ABSA at the token level, which exploits aspect and corresponding polarity at the span level. However, this dataset has not mined the rationale of the holder and target of the polarity in the sentence toward the aspects. Hence, our motivation is to introduce ViTOED, which exploits emotion and expression by targeting tokens at the input sentence level. 

\section{Dataset}
\label{dataset}

\subsection{Data Creation Process}


We use data from social networks, including 6,000 sentences from the UIT-VSMEC \cite{ho2020emotion}, and 5,010 comments were collected raw from our social media platforms.  At first, we give the annotation guidelines to the annotators. Then, we let the annotators annotate a set of 210 sample sentences to evaluate the agreement among annotators. If the agreement between annotators is adequate, we let them annotate the entire dataset. After that, we perform cross-checking among annotators to ensure accurate labeling. To make labeling more convenient for annotators, we use Doccano\footnote{\url{https://github.com/doccano/doccano}} to annotate the data with a friendly user interface and easy setup on individual devices. 

\subsection{Annotation Guidelines}

For each sentence, the annotator must annotate the quadruple: source, target, expression, and polarity. There may be more than one quadruple per sentence. The meanings of source, target, expression, and polarity are described below:

\begin{table}[ht]
    \small
    \centering
    \caption{Examples from the ViTOED dataset}
    \resizebox{.48\textwidth}{!}{
    \begin{tabular}{c|p{5cm}|p{5cm}}
    \textbf{No.} & \textbf{Comment} & \textbf{Label} \\
    \hline
    1 & \makecell[lp{5cm}]{\textcolor{blue}{tao} \textcolor{cyan}{thích} \textcolor{red}{bài này} lắm luôn ý \\ (\textbf{English}:  \textcolor{blue}{I} \textcolor{cyan}{like} \textcolor{red}{this song} very much)} & \makecell[lp{5cm}]{
        \textbf{Source}: \textcolor{blue}{“tao”}, \\ 
        \textbf{Target}: \textcolor{red}{“bài này”}, \\ 
        \textbf{Expression}: \textcolor{cyan}{“thích”}, \\ 
        \textbf{Polarity}: \textit{Positive}
    } \\
    \hline
    2 & \makecell[lp{5cm}]{\textcolor{red}{7 thằng này} \textcolor{cyan}{không phải con người chúng mày ạ} \\ (\textbf{English}: \textcolor{red}{These 7 guys} \textcolor{cyan}{are not human})} & \makecell[lp{5cm}]{
        \textbf{Source}: “”, \\ 
        \textbf{Target}: \textcolor{red}{“7 thằng này”}, \\ 
        \textbf{Expression}: \textcolor{cyan}{“không phải con người chúng mày ạ”}, \\ 
        \textbf{Polarity}: \textit{Positive}
    } \\
    \hline
    3 & \makecell[lp{5cm}]{\textcolor{blue}{tao} \textcolor{cyan}{ho rụng hết tóc rồi} \\ (\textbf{English}: \textcolor{blue}{I} \textcolor{cyan}{coughed so much that my hair fell out})} & \makecell[lp{5cm}]{
        \textbf{Source}: \textcolor{blue}{“tao”}, \\ 
        \textbf{Target}: “”, \\ 
        \textbf{Expression}: \textcolor{cyan}{“ho rụng hết tóc rồi”}, \\ 
        \textbf{Polarity}: \textit{Negative}
    } \\
    \hline
    \end{tabular}
    }
    
    \label{fig_example}
\end{table}

\textbf{Source}: The origin of the comment comes from the sender, including an individual or organization. The source of comments expresses an individual's views, feelings, and actions towards the object, thing, or phenomenon mentioned. The source of sentiment is subjective to the commentator and is revealed through many different emotional forms. Aspects often originate from nouns and pronouns and are usually in the first person (i.e., "tao" in Vietnamese).

\textbf{Target}: The target refers to an individual, group, object, or phenomenon, which is directed by the "speaker" (the source) to express emotions, states, thoughts, etc. that affect the "object" (the target). The subjects are derived from nouns, pronouns, and characters. They are usually in the second person (For example, "mày", "bài này", "7 thằng này" as shown in Table \ref{fig_example}). In addition, the target can contain additional noun phrases of ownership relations. 

\textbf{Expression}: Expressing the source's emotions, opinion, feelings, and actions toward a particular object or event. Emotions include adjectives that describe emotions, or words that have an exclamative meaning (exclamation words), i.e., "đáng sợ lắm", or action (verb) accompanied by interjections and adjectives describing emotions, i.e., "ho rụng hết tóc rồi", as shown in Table \ref{fig_example}.

\textbf{Polarity}: Describe the negativity, positivity, and neutrality in the expression, which comprises three categories:  \textit{Positive} expresses the optimistic, encouraging, motivating, and empathetic emotion, \textit{Negative} expresses offense, incite, hate, and scare contents, and \textit{Neutral} indicates normal expressions that are not positive or negative. In addition, humor expressions that do not contain hateful or offensive content can be classified as \textit{Positive} while the expressions that are untrue and misleading, as well as expressions that contain profanity, rudeness, and sensitive aspects, are also classified as \textit{Negative}. 

\subsection{Inter Annotator Agreement Evaluation}

\begin{table*}[ht!]
    \centering
    \caption{Overview statistic about the ViTOED dataset.}
    \label{tab_overview}
    \resizebox{.8\textwidth}{!}{
    \begin{tabular}{@{}lrrrrrrrrrrrrrr@{}}
    \toprule
    \textbf{} & \multicolumn{2}{c}{\textbf{Texts}} & \multicolumn{3}{c}{\textbf{Source}} & \multicolumn{3}{c}{\textbf{Target}} & \multicolumn{3}{c}{\textbf{Expression}} & \multicolumn{3}{c}{\textbf{Polarity}} \\
    \cmidrule(lr){2-3} \cmidrule(lr){4-6} \cmidrule(lr){7-9} \cmidrule(lr){10-12} \cmidrule(lr){13-15}
    & \# & avg. & \# & avg. & max & \# & avg. & max & \# & avg. & max & negative & neutral & positive \\
    \cmidrule(lr){2-3} \cmidrule(lr){4-6} \cmidrule(lr){7-9} \cmidrule(lr){10-12} \cmidrule(lr){13-15}
    train & 7,706 & 16.9 & 2,359 & 1.4 & 8 & 8,632 & 3.0 & 20 & 15,024 & 6.5 & 29 & 6,571 & 3,936 & 4,517 \\
    dev & 1,085 & 16.0 & 385 & 1.7 & 12 & 1,160 & 3.1 & 16 & 2,087 & 7.0 & 21 & 962 & 505 & 620  \\
    test & 2,194 & 16.8 & 572 & 1.2 & 6 & 2,127 & 2.8 & 20 & 4,133 & 6.9 & 24 & 1,681 & 1,366 & 1,086 \\
    \bottomrule
    \end{tabular}
    }
\end{table*}

To measure Inter-Annotator Agreement (IAA) among annotators, we need to calculate the span-based intersection of the Source, Target, and Expression entities. Following the work by \cite{thanh-etal-2021-span}, we employ the F1-score for measuring the agreement among annotators. The F1-Score is computed as shown in Equation \ref{eq_f1}, where $a$ and $b$ are the first and second annotators, $A$ and $B$ are the set of tokens annotated by the first and second annotators:
\begin{equation} 
    \label{eq_f1}
    F_1(a, b) = \frac{2 \cdot |A \cap B|}{|A| + |B|}
\end{equation}

For Polarity, we must retrieve the opinion quadruples and then use pairwise to calculate the number of overlapping labels for the two annotators, dividing by the corresponding number of opinion quadruples for each annotator. Finally, we take the average overlapping pairwise between two annotators for the final agreement in Polarity, as shown in Equation \ref{eq_pair_final}, where a and b are the first and second annotators, A and B are the set (S, T, E, P) of matched labels between two annotators. 




\begin{equation} 
    \label{eq_pair_final}
P = \frac{P(a, b) + P(b, a)}{2}=\frac{\frac{|A \cap B|}{|A|}+\frac{|B \cap A|}{|B|}}{2}
\end{equation}

\begin{table}[ht]
    \centering
    \caption{Inter-rater agreement degree.}
    \label{tab_results}
    \resizebox{.4\textwidth}{!}{
    \begin{tabular}{@{}ccccc@{}}
    \hline
    \textbf{Round} & \textbf{Source} & \textbf{Target} & \textbf{Expression} & \textbf{Polarity} \\ \midrule
    1 & 0.17 & 0.31 & 0.34 & 0.15 \\ 
    2 & 0.26 & 0.48 & 0.54 & 0.35 \\ 
    3 & 0.27 & 0.51 & 0.64 & 0.40 \\ 
    4 & \textbf{0.32} & \textbf{0.61} & \textbf{0.77} & \textbf{0.47} \\ \midrule
    \end{tabular}
    }
\end{table}


Since we have a group of 3 annotators, the final IAA is computed by averaging across the pairs. To ensure that annotators understand the guidelines and make accurate annotations, we train them through multiple rounds. Table \ref{tab_results} describes the inter-annotator agreement result between annotators after four rounds. It can be seen that the agreement level improves after each round, indicating the efficiency of the annotation process and the guidelines. The Target and Expression show high IAA after four rounds, indicating strong consensus among annotators. On the other hand, as described in Table \ref{tab_results}, the low IAA for the Source is because many sentences are difficult to detect as containing a source (Examples \#3 and \#4 in Table \ref{fig_example}), which leads to annotators being confused, resulting in omissions or errors, making it challenging for the annotators to identify and label them accurately. For "Polarity," it depends on the consistency across (Source, Target, Expression, Polarity) tuples, so the agreement rate of approximately 0.5 is moderate. The level of agreement indicates that the annotators understood the labeling guidelines and followed the instructions accurately. 

\subsection{Dataset Analysis}
The ViTOED datasets consist of 21,244 annotated opinion quadruples within 10,985 sentences from social media networks. We divided the dataset into three subsets: Train, Dev, and Test, with a ratio of 7:1:2. This was done to facilitate data analysis, observation, as well as model training and parameter tuning. Table \ref{tab_overview} provides an overview of the ViTOED dataset. The \textit{max} and \textit{avg} are computed at the word level. In general, the distributions of source, target, and expression across the training, development, and test sets are similar. The sentiment tends to lean slightly toward the negative class. However, the differences among the three polarities are not significant.


Although the number of sources and targets is relatively small compared to the number of opinions, their distribution is appropriately aligned with the proportions in the train, dev, and test data splits. We can also observe that the number of targets is only half the number of opinions, indicating that targets are generally mentioned at an average frequency in social media comments. As for the sources, given the nature of social media comments and the structure of Vietnamese, sources are often implicit or assumed to be the commenters, resulting in a relatively low number of explicit sources. In Table \ref{tab_opinion_category}, we define four types of opinions in the comments of the dataset. The T-E demonstrated that an opinion missing the Source accounts for 52.10\% of the total opinions. This type of opinion is the most common in social media comments today. 

\begin{table}[H]
    \centering
    \caption{\centering Dataset Distribution by Opinion Category. \textbf{S} is the Source, \textbf{T} is the Target, and \textbf{E} is the Expression}
    \label{tab_opinion_category}
    \resizebox{.32\textwidth}{!}{
    \begin{tabular}{@{}lcccc@{}}
    \toprule
    \textbf{Entity}     & \textbf{S-T-E} & \textbf{S-E} & \textbf{T-E} & \textbf{E} \\ \midrule
    \textbf{Train}      & 1,667 & 1,064 & 8,054 & 4,239 \\ \midrule
    \textbf{Dev}        & 299   & 329   & 1,922 & 1583   \\ \midrule
    \textbf{Test}       & 277   & 134   & 1,093 & 583 \\ \midrule
    \textbf{Total}      & 2,243 & 1,527 & 11,069 & 6,405 \\ \bottomrule
    \end{tabular}
    }
\end{table}

\begin{table*}
    \centering
    \caption{Top words for SOURCE, TARGET, and EXPRESSION}
    \label{tab:top_words}
    \resizebox{.8\textwidth}{!}{
    \begin{tabular}{@{}ccccccc|cc|cc@{}}
    \toprule
    \textbf{} & \multicolumn{2}{c}{\textbf{SOURCE}} & \multicolumn{2}{c}{\textbf{TARGET}} & \multicolumn{6}{c}{\textbf{EXPRESSION}} \\ 
    \cmidrule(lr){2-3} \cmidrule(lr){4-5} \cmidrule(lr){6-11}
    \textbf{} & \textbf{} & \textbf{} & \textbf{} & \textbf{} & \multicolumn{2}{c}{\textbf{POSITIVE}}& \multicolumn{2}{c}{\textbf{NEUTRAL}}& \multicolumn{2}{c}{\textbf{NEGATIVE}} \\
    \cmidrule(lr){2-3} \cmidrule(lr){4-5} \cmidrule(lr){6-11}
    \textbf{TOP} & \textbf{Word} & \textbf{Freq} & \textbf{Word} & \textbf{Freq} & \textbf{Word} & \textbf{Freq} & \textbf{Word} & \textbf{Freq} \ & \textbf{Word} & \textbf{Freq} \\ 
    \midrule
    1 & tao (I)  & 881 & mày (You) & 396 & yêu (Love) & 76 & đi (Go) & 39 & sợ (Scare) & 80 \\
    2 & t (I) & 174 & mấy (any) & 374 & thương (Love) & 66 & ko (Not) & 19 & đéo (f*ck) & 52 \\
    3 & ta (I) & 90 & thằng (he) & 349 & đẹp (beauty) & 47 & mua (buy) & 14 & buồn (sad) & 41 \\
    4 & e (younger) & 81 & tao (me) & 230 & vui (happy) & 38 & bảo (said) & 14 & chửi (scold) & 40\\
    5 & Em (younger) & 66 & ta (me) & 204 & cảm ơn (thanks) & 35 & coi (see) & 13 & ghét (hat) & 35\\
    \bottomrule
    \end{tabular}
    }
\end{table*}

Additionally, we investigate the frequency of words in each component —Source, Target, and Expression. We select the top-5 most frequently appearing words, as shown in Table \ref{tab:top_words}. For the source, most of the words are Vietnamese personal pronouns, i.e., "tao", "em", "mình", "tui", and "ta" (in English, these personal pronouns are I). In Vietnamese, the personal pronouns "tao", "em", "mình", "tui", and "ta" are commonly used to refer to individuals in communication. Moreover, users tend to use the abbreviated forms of personal pronouns on social media, such as "t" for "tao" and "e" for "em". Overall, the "tao" (in English, similar to "I") and its abbreviated form "t" are mainly used to make the source. For the target, the popular personal pronouns are "mày", "thằng", "bọn", and "đứa" (in English, these personal pronouns are "You"). In addition, the word "mấy" also appeared frequently in the target. This word serves as an adjunct that takes a plural form of personal pronouns in Vietnamese. For example, the word "mấy thằng" means a group of people. Especially, according to Table \ref{tab:top_words}, the words "mày" (English: you) and "tao" (English: I) are used frequently to express both source and target in the sentence, although the "tao" mostly expresses the source, while "mày" focuses on describing the target. In Vietnamese, these words are commonly used in communication between friends or groups of people who are acquainted.

Finally, we investigate the word frequency for the expression according to the three polarities as shown in Table \ref{tab:top_words}. Based on each polarity, there are several characteristics of words such as "yêu", "thuơng" (English: love), and "đẹp" (English: beautiful) that express the positive polarity, "sợ" (English: scare), "đéo" (English: f*ck), and "buồn" (English: sad) express for the negative emotion, and "đi" (English: let's or go), "ko" (English: no, don't) shows the neutral expression. 

\section{Methodology}
\label{method}

\begin{figure}[ht!]
    \centering
    \includegraphics[width=.5\textwidth]{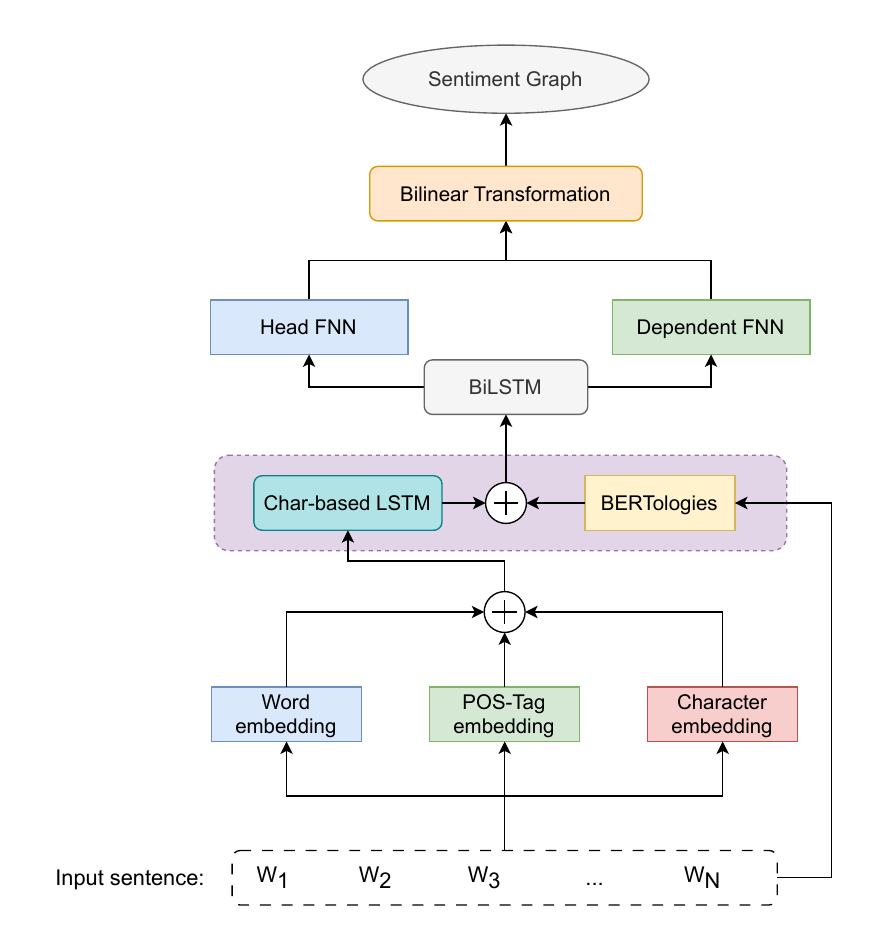}
    \caption{The sentiment graph baseline model.}
    \label{fig:4}
\end{figure}

We employ the Structured Sentiment Graph proposed by \cite{barnes-etal-2021-structured} as our baseline model. Figure \ref{fig:4} illustrates the overview of the baseline model. First, we tokenized the input sentences into token level using the spaCy tools for Vietnamese\footnote{\url{https://github.com/trungtv/vi_spacy}}. Then, the sequence of tokenized tokens is passed through the word, pos-tag, and character embedding layers, and then through an LSTM layer to construct character representation vectors. These vectors are concatenated to generate a contextualized embedding that captures the sentence's semantic meaning at the token level.

Additionally, the input sentence is passed through a BERT model to generate a BERT-contextualized vector, enhancing token representations. Then, the character-based LSTM and ERT-contextualized vectors are passed into a BiLSTM to create the contextualized representation. Finally, we create a sentiment graph by fitting the contextualized embedding into two Feed-forward networks (FNN), including the HeadFNN and DependentFNN.

\begin{figure}[ht]
    \centering
    \label{fig_5}
    \begin{subfigure}[b]{.5\textwidth}
        \centering
        \includegraphics[width=\textwidth]{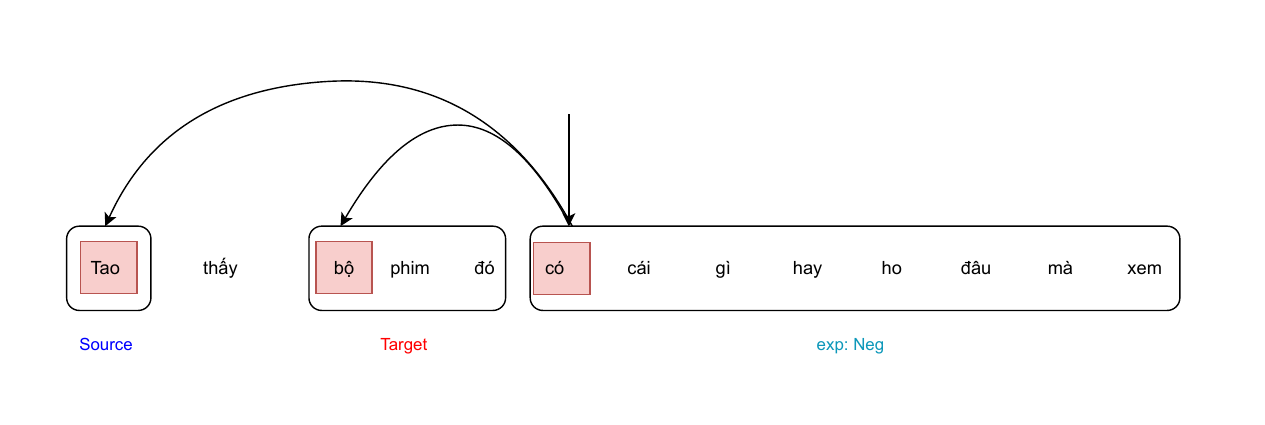}
        \caption{head-first}
        \label{fig_5_a}
    \end{subfigure}
    ~
    \begin{subfigure}[b]{.5\textwidth}
        \centering
        \includegraphics[width=\textwidth]{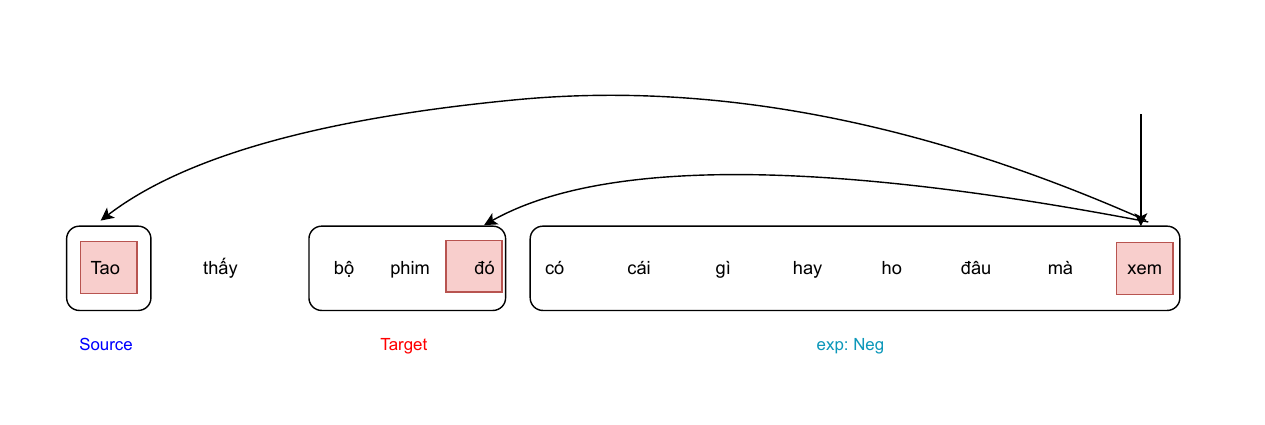}
        \caption{head-final}
        \label{fig_5_b}
    \end{subfigure}
    ~
    \caption{The example of the parsing graph representation.
    (Vietnamese: \textcolor{blue}{Tao} thấy \textcolor{red}{bộ phim đó} \textcolor{cyan}{có cái gì hay ho đâu mà xem}. English: \textcolor{blue}{I} think \textcolor{red}{that movie} \textcolor{cyan}{isn’t worth watching.}) }
    \label{fig_app_graph}
\end{figure}

To represent the sentiment graph, we use the two parsing graph representations by \cite{barnes-etal-2021-structured}, including the \textbf{head-first} and \textbf{head-final} nodes. The \textit{head-first} sets the first token in the span as the root and other tokens within the span as the dependents (as illustrated in Figure \ref{fig_5_a}). On the other hand, the \textit{head-final} sets the final token in the span as the root, and other tokens in the spans as dependents (as shown in Figure \ref{fig_5_b}).

Finally, we employ two types of BERTology models for Vietnamese —multilingual and monolingual —to represent contextual embeddings. The multilingual models comprise m-BERT \cite{devlin-etal-2019-bert} and XLM-R \cite{conneau-etal-2020-unsupervised}, while the monolingual models in the Vietnamese language consist of PhoBERT \cite{nguyen-tuan-nguyen-2020-phobert}, ViSoBERT \cite{nguyen-etal-2023-visobert}, and CafeBERT \cite{do-etal-2024-vlue}. Additionally, we also employ two Vietnamese encoder-decoder models, including ViT5 \cite{phan-etal-2022-vit5} and BARTPho \cite{tran2021bartpho}, and multilingual models supporting Vietnamese, including mT5 \cite{xue-etal-2021-mt5} and mBART \cite{liu-etal-2020-multilingual-denoising}.

\section{Experimental Results}
\label{results}
\subsection{Experimental results}

To evaluate the performance of the baseline model, we follow four metrics as introduced in \cite{barnes-etal-2021-structured}, including Span F1, Targeted F1, Parsing Graph F1(UF1 and LF1 regarding unlabeled and labeled arcs of the parsing graph), and Sentiment Graph F1 (NSF1 and SF1 regarding Non-polar Sentiment Graph and Sentiment Graph). Table \ref{tab:comparison} illustrates the results of the performance of baseline models employed by different Vietnamese pre-trained language models. 

\begin{table}[ht]
    \centering
    \caption{Performance comparison of different models across various metrics on the test set. The bold indicates the best on head-first, and the underline demonstrates the best on head-final.}
    \label{tab:comparison}
    \resizebox{.49\textwidth}{!}{
    \begin{tabular}{@{}cccccccccc@{}}
    \toprule
    \textbf{} & \textbf{} & \multicolumn{3}{c}{\textbf{Spans}} & \textbf{Targeted} & \multicolumn{2}{c}{\textbf{Parsing Graphs}} & \multicolumn{2}{c}{\textbf{Sent. Graphs}} \\ 
    \cmidrule(lr){3-5} \cmidrule(lr){6-6} \cmidrule(lr){7-8} \cmidrule(lr){9-10}
    \textbf{} & \textbf{} & \textbf{Source} & \textbf{Target} & \textbf{Exp} & \textbf{F1} & \textbf{UF1} & \textbf{LF1} & \textbf{NSF1} & \textbf{SF1} \\ 
    \midrule
    \multirow{2}{*}{\textbf{mBERT}} &  head-first &  57.07 &    56.30 & 74.57 &         28.08 & 55.46 & 35.73 & 50.46 & 31.20 \\
                                         & head-final & 53.87 &    54.82 & 77.21 &         22.68 & 62.73 & 37.97 & 49.23 & 29.34 \\
                                         \cmidrule(lr){2-10}
        \multirow{2}{*}{\textbf{XLM-R}} &  head-first & 57.50 &    56.04 & 69.43 &         27.50 & 55.00 & 34.15 & 51.09 & 29.89 \\
                                          & head-final & 52.73 &    54.45 & 77.49 &         23.82 & 64.15 & 40.05 & 50.75 & 30.63 \\
                                          \cmidrule(lr){2-10} 
        \multirow{2}{*}{\textbf{mBART}} &  head-first & 55.46 &    54.46 & 65.36 &         26.46 & 52.97 & 33.01 & 48.07 & 28.17 \\
                                        & head-final & 55.44 &    51.93 & 74.63 &         21.15 & 62.73 & 37.21 & 49.81 & 29.02 \\
                                        \cmidrule(lr){2-10}
        \multirow{2}{*}{\textbf{mT5}} &  head-first & \textbf{60.00} &    56.53 & 71.14 &         28.90 & 55.96 & 35.38 & 50.42 & 30.64 \\
                                        & head-final & 53.65 &    52.20 & \underline{80.78} &         22.11 & 66.52 & 39.04 & \underline{52.32} & 29.41 \\
                                        \cmidrule(lr){2-10}
        \multirow{2}{*}{\textbf{ViSoBERT}}& head-first & 59.14 &    56.88 & 72.70 &  \textbf{30.80} & 56.39 & 36.85 & 52.66 & 33.34 \\
                                          & head-final & 54.18 &    \underline{55.62} & 77.39 &         \underline{25.82} & 63.72 & 40.55 & 51.20 & 33.07 \\
                                          \cmidrule(lr){2-10}
        \multirow{2}{*}{\textbf{PhoBERT}} & head-first & 59.03 &    56.94 & 74.13 &         27.04 & 55.21 & 33.54 & 52.88 & 29.02 \\
                                          & head-final & 54.80 &    55.12 & 78.71 &         23.39 & 65.15 & 39.88 & 51.01 & 30.88 \\ 
                                          \cmidrule(lr){2-10}
        \multirow{2}{*}{\textbf{CafeBERT}} & head-first & 58.12 &    \textbf{56.96} & 74.56 &         29.24 & 57.05 & 36.00 & 51.20 & 31.84 \\
                                         & head-final & 52.99 &    53.69 & 73.34 &         24.14 & 61.52 & 38.69 & 49.68 & 31.03 \\
                                         \cmidrule(lr){2-10}
        \multirow{2}{*}{\textbf{ViT5}} & head-first & 59.62 &    55.44 & \textbf{75.61} &         29.91 & \textbf{57.61} & \textbf{38.41} & \textbf{53.86} & \textbf{33.97} \\
        & head-final & 54.65 &    54.96 & 80.56 &         24.69 & 65.84 & \underline{42.40} & 51.67 & \underline{33.40} \\
        \cmidrule(lr){2-10}
        \multirow{2}{*}{\textbf{BARTPho}} & head-first & 52.18 &    52.90 & 52.96 &         18.56 & 48.22 & 26.92 & 44.04 & 19.97 \\
        & head-final & 41.33 &    45.07 & 68.10 &         11.34 & 60.89 & 29.97 & 45.87 & 20.43 \\
        \bottomrule
        \end{tabular}
    }
\end{table}

According to Table \ref{tab:comparison}, the monolingual pre-trained language models, such as ViSoBERT and PhoBERT achieve better performance than multilingual models, such as mBERT and XLM-R, on this task. Besides, the head-first approach is efficient for extracting entities in a sentence, such as the source and target, whereas the head-final approach is suitable for detecting expressions. For the parsing graph, the head-final approach also performs better than the head-first approach. In contrast, the head-first and head-final approaches obtain similar performance in representing the sentiment graph. Additionally, mT5 and CafeBERT achieved the highest results for the target-oriented emotion detection task, in which they are good at extracting the source and target entities, respectively, while ViT5 is robust at exploiting the expression, parsing, and sentiment graphs, and ViSoBERT shows high performance in detecting the targeted expression. 

\subsection{Error Analysis}

The cause of label misalignment is the model's failure to predict missing entities or to correctly align entities. 
The model has not yet handled spans well and is prone to errors across spans, resulting in lower performance. As shown in Table \ref{tab:graph_annotate_result}, the target entity is misaligned across two spans, and the model also failed to predict the Expression, resulting in a decrease in accuracy. The inaccurate predictions are caused by the model confusing the Source and Target, as both labels are associated with person-related or possessive nouns referring to people. This confusion between the two labels is unavoidable, and the language model has not yet handled this type of noun effectively 

\begin{table}[ht]
    \centering
    \caption{Results of graph annotation by opinion components by F1(\%).}
    \label{tab:graph_annotate_result}
    \resizebox{.4\textwidth}{!}{
    \begin{tabular}{lrrrrr}
        \toprule
        & \multicolumn{1}{c}{\multirow{2}{*}{\textbf{Source}}} & \multicolumn{1}{c}{\multirow{2}{*}{\textbf{Target}}} & \multicolumn{3}{c}{\textbf{Expression}}  \\
        \cmidrule(lr){4-6}
        & \multicolumn{1}{c}{}                                 & \multicolumn{1}{c}{}                                 & \multicolumn{1}{c}{\textbf{negative}} & \multicolumn{1}{c}{\textbf{neutral}} & \multicolumn{1}{c}{\textbf{positive}} \\
        \midrule
    \textit{head-first}         & 29.95                                                & 44.92                                                & 43.77                                 & 19.87                                & 43.77                                 \\
    \textit{head-first+inlabel} & 40.94                                                & 39.69                                                & 48.90                                 & 33.08                                & 46.35                                 \\
    \textit{head-final}         & 24.35                                                & 40.89                                                & 55.74                                 & 22.41                                & 34.24                                 \\
    \textit{head-final+inlabel} & 32.51                                                & 36.35                                                & 62.01                                 & 25.48                                & 36.77                                 \\
    \textit{Dep. edge}          & 26.14                                                & 16.23                                                & 32.54                                 & 14.73                                 & 25.42                                 \\
    \textit{Dep. label}         & 25.65                                                & 14.53                                                & 28.88                                 & 11.66                                 & 27.36      \\
    \bottomrule
    \end{tabular}
    }
\end{table}



\begin{figure}[ht!]
    \centering
    \begin{subfigure}[b]{.5\textwidth}
        \centering
        \includegraphics[width=\textwidth]{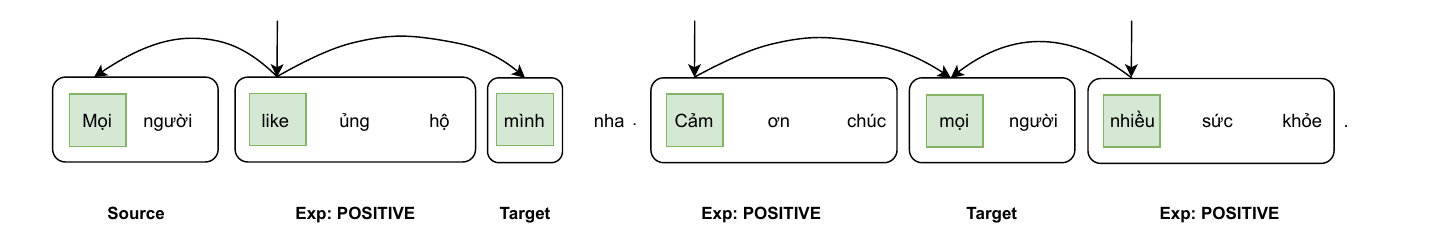}
        \caption{Ground truth}
        \label{fig:error3_a}
    \end{subfigure}
    ~
    \begin{subfigure}[b]{.5\textwidth}
        \centering
        \includegraphics[width=\textwidth]{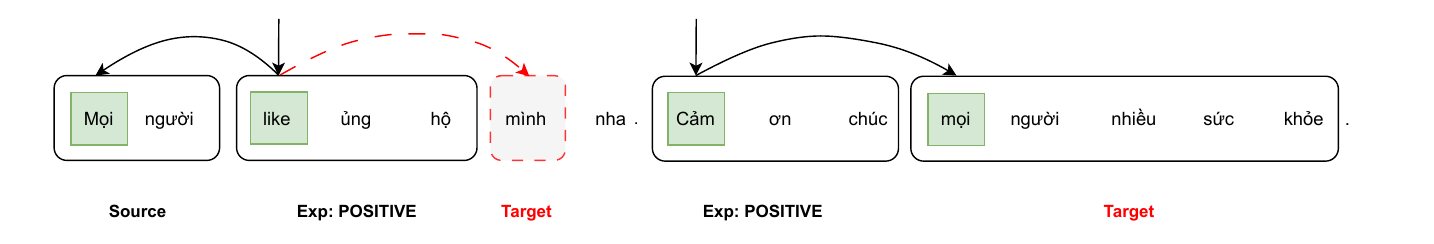}
        \caption{Prediction}
        \label{fig:error3_b}
    \end{subfigure}
    \caption{Error sample: the edges are misaligned between tokens. (\textit{English}: Please support me by liking this. Thank you, and wishing everyone good health.)}
    \label{fig:error3}
\end{figure}

In addition, we also test the \textit{+inlabel} structure that adds information to nodes that are not the main head in a graph. The \textit{Dep. edges} uses syntactic information to identify the main head of each span while \textit{Dep. labels} is a filtered version of \textit{Dep. edges} to remove unreasonable choices, such as punctuation or peripheral words, when selecting the head. According to Table \ref{tab:graph_annotate_result}, adding the +inlabel structure improved the accuracy in identifying and extracting these spans. (In Table \ref{tab:graph_annotate_result}, the results are recorded on the development set by the ViT5 model). Besides, the incorrect predictions are due to the model's weak handling of edges and roots. When using graphs with the head-first or head-final approach, the root is either the starting or ending point of the token, and the remaining tokens in the same set should be placed along the edges. However, the model mistakenly places the root, leading to misaligned token edges and label prediction errors (see the error example in Figure \ref{fig:error3}).

To sum up, based on the error analysis, we identified three main challenges for target-oriented emotion detection, as reflected in the baseline's performance on the ViTOED. First, there is difficulty in recognizing source, target, and expression entity spans in the text due to the complexity of the sentence vocabulary. Second, the misprediction between the root and edges makes it challenging to detect the correct target emotion expressed by the subject from the sentence. Finally, confusion between source and target due to Vietnamese morphology can affect model performance in correctly identifying the source and target in a sentence.


\section{Conclusion}
\label{conclusion}
This paper presents ViTOED, a human-annotated dataset for target-oriented emotion detection in Vietnamese social media texts. ViTOED consists of 21,244 annotated opinions from 10,985 comments of users on social networks. Besides, we construct a baseline model based on the structured sentiment graph architecture \cite{barnes-etal-2021-structured} and evaluate various Vietnamese Pre-trained language models on the dataset. From the error analysis, the model still exhibits prediction errors, including span misalignment between source and target, misprediction of relations, and confusion between target and source entities, indicating a challenge for further improvement in language understanding and for correctly mining sentiments. The source code and dataset are available at: \url{https://github.com/sonlam1102/vitoed}.


\section*{Acknowledgments}

This research is funded by Vietnam National University HoChiMinh City (VNU-HCM) under grant number NCM2025-26-02

\bibliographystyle{IEEEtran}
\bibliography{references}

\end{document}